\pdfoutput=1

\documentclass[11pt]{article}

\usepackage[final]{acl}

\usepackage{times}
\usepackage{latexsym}

\usepackage[T1]{fontenc}

\usepackage[utf8]{inputenc}

\usepackage{microtype}

\usepackage{inconsolata}

\usepackage{graphicx}
\usepackage{booktabs}
\usepackage{leipzig}
\usepackage{gb4e}
\usepackage{subcaption}
\usepackage{multirow}
\noautomath
\makeglossaries

\newleipzig{npst}{npst}{nonpast}
\newleipzig{qp}{qp}{questionparticle}

\newcommand{\dataset}{SGET}

\title{Evaluating Structural Generalization in Neural Machine Translation}

\author{Ryoma Kumon \quad Daiki Matsuoka \quad Hitomi Yanaka \\
  The University of Tokyo \\
  \texttt{\{kumoryo9, daiki.matsuoka, hyanaka\}@is.s.u-tokyo.ac.jp}}

\begin{document}
\maketitle
\begin{abstract}
Compositional generalization refers to the ability to generalize to novel combinations of previously observed words and syntactic structures.
Since it is regarded as a desired property of neural models, recent work has assessed compositional generalization in machine translation as well as semantic parsing.
However, previous evaluations with machine translation have focused mostly on lexical generalization (i.e., generalization to unseen combinations of known words).
Thus, it remains unclear to what extent models can translate sentences that require structural generalization (i.e., generalization to different sorts of syntactic structures).
To address this question, we construct \dataset{}, a machine translation dataset covering various types of compositional generalization with control of words and sentence structures.
We evaluate neural machine translation models on \dataset{} and show that they struggle more in structural generalization than in lexical generalization.
We also find different performance trends in semantic parsing and machine translation, which indicates the importance of evaluations across various tasks.
\end{abstract}
\section{Introduction}
\label{sec:introduction}
Humans can understand and produce novel language expressions by combining familiar words and syntactic structures~\citep{partee1984compositionality, fodor1988connectionism}.
This ability is known as \emph{compositional generalization}, which has recently attracted much attention in the context of investigating the generalization ability of neural models~\citep{lake-2018-scan}.
Compositional generalization is divided into two categories: lexical generalization and structural generalization~\citep{kim-linzen-2020-cogs, yao-koller-2022-structural, li-etal-2023-slog}.
Lexical generalization is generalization to an unseen combination of familiar lexical items regardless of syntactic structures, whereas structural generalization is generalization to an unseen combination of familiar syntactic structures and lexical items.
\begin{figure}[t]
    \centering
    \begin{minipage}[b]{\linewidth}
    \includegraphics[width=\linewidth]{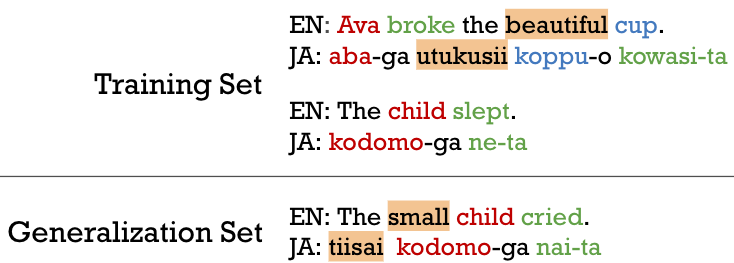}
    \subcaption{English-Japanese translation}
    \end{minipage}
    \begin{minipage}[b]{\linewidth}
    \includegraphics[width=\linewidth]{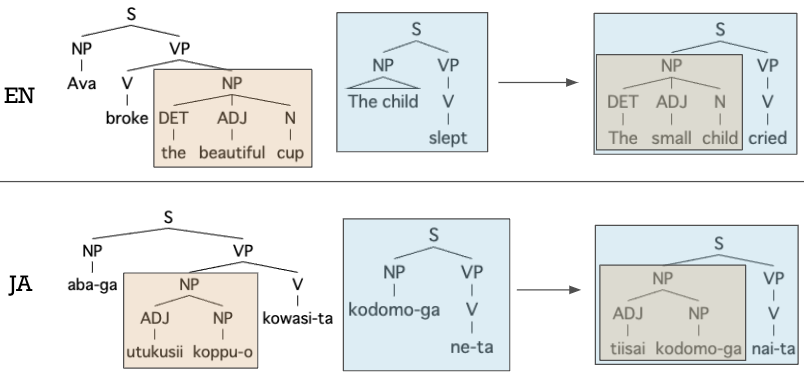}
    \subcaption{Syntactic trees}
    \end{minipage}
    \caption{How we test the compositional generalization abilities of neural models on \dataset{}.
    This pattern focuses on the structural generalization to an adjective modifying a subject noun, which is unseen in the training.}
    \label{fig:fig1}
\end{figure}

Existing studies have investigated compositional generalization mainly with semantic parsing~\citep{kim-linzen-2020-cogs, li-etal-2023-slog}, considering both lexical and structural generalization.
On the other hand, evaluation with application tasks is also essential to thoroughly evaluate the compositional generalization abilities of neural models~\citep{yao-koller-2022-structural}.
Among those application tasks, several studies have focused on machine translation~\citep{li-etal-2021-compositional, dankers-etal-2022-paradox, moisio-etal-2023-using}, as it can be considered a mapping from a source language to a target one.
The translation result by a model should indicate how the model handles the lexical and syntactic features because correct translation requires a model to reflect the lexical and syntactic features of the source sentence.
However, these studies focused primarily on lexical generalization, and structural generalization has yet to be investigated due to the difficulty of precise control of structural gaps between source and target sentences.

In this paper, we propose \dataset{}\footnote{\url{https://github.com/ynklab/SGET}} (\textbf{S}tructural \textbf{GE}neralization Benchmark based on English-Japanese Machine \textbf{T}ranslation), a parallel dataset for English-Japanese translation tasks covering both lexical and structural generalization.
Figure~\ref{fig:fig1} shows an example of a structural gap between the training and generalization sets in \dataset{}.
The training set includes sentences where the object is a noun phrase in which an adjective modifies a noun.
If a model generalizes compositionally, then it should be able to translate sentences with a noun phrase of the same structure in the subject position.

We adopt a rule-based method to construct \dataset{} so that we can control lexical items and syntactic structures and strictly evaluate the compositional generalization abilities of models.
The method involves generating English sentences with Probabilistic Context-Free Grammar (PCFG) and creating Japanese parallel translations with rule-based machine translation (RBMT).
We design the experimental settings so that non-essential factors such as sentence length~\citep{wu-etal-2023-recogs} and naturalness do not affect the evaluation.
We evaluate LSTM~\citep{hochreiter1997long}, Transformer~\citep{vaswani-2017-transformer}, and Llama~2~\citep{touvron2023llama} on \dataset{} and analyze their performances.
We also compare the results with those of previous studies based on semantic parsing.

Our contributions are as follows.
\begin{enumerate}
    \item We propose a rule-based method for constructing a parallel dataset for translation tasks in a controlled manner.
    \item Using the proposed method, we introduce \dataset{}, which covers lexical and structural generalization patterns.
    \item We assess the compositional generalization abilities of neural models on \dataset{}.
\end{enumerate}

\section{Related Work}
\label{sec:background}
\subsection{Semantic Parsing}
\label{subssec:semantic-parsing}
Various methods and benchmarks in semantic parsing have been proposed to evaluate the compositional generalization capacity of neural models.
\citet{lake-2018-scan} introduced SCAN, which presents the task of converting commands generated with limited vocabulary and grammar into action sequences.
\citet{kim-linzen-2020-cogs} proposed COGS, the task to map sentences generated with more diverse vocabulary and grammar than in SCAN to semantic representations.
\citet{kim-linzen-2020-cogs} investigated the performances of 
LSTM and Transformer on lexical generalization and structural generalization and showed that the latter poses more challenges.
Our work also focuses on both lexical and structural generalization in machine translation, and we discuss the difference between the model performance on COGS (semantic parsing) and that on \dataset{} (machine translation) in Section~\ref{sec:task}.

Some studies have worked on improving COGS and validating its experimental settings.
\citet{li-etal-2023-slog} proposed SLOG, a dataset that expanded the structural generalization patterns in COGS to test structural generalization more thoroughly.
\citet{wu-etal-2023-recogs} mitigated the issue of COGS involving semantically non-essential factors of meaning representations, such as their length and variable binding.
\citet{csordas-etal-2021-devil} showed that the compositional generalization abilities of the models were underestimated in the experimental settings of COGS.
They also observed that higher accuracy can be achieved by using relative positional encoding and turning off early stopping.
We incorporate these modifications in \dataset{} and analyze their impact on model performance.

\citet{wang-hershcovich-2023-evaluating} focused on cross-lingual compositional generalization in semantic parsing.
They showed that neural machine translation is inconsistent in terms of lexical and syntactic aspects and that RBMT is a better translation method to provide a cross-lingual benchmark for compositional generalization.

\citet{yao-koller-2022-structural} argued that evaluations based on semantic parsing do not fully reveal the compositional generalization abilities of models because outputs in semantic parsing are logical formulas, not natural language sentences.
They adopted question-answering, an application task whose output is natural language, to test structural generalization in a text-to-text format.

\subsection{Machine Translation}
\label{subsec:machine-translation}
Several studies~\citep{raunak2019compositionality, li-etal-2021-compositional, dankers-etal-2022-paradox, shi-etal-2022-revisit, moisio-etal-2023-using, zheng-lapata-2023-real} evaluated compositional generalization in machine translation, which is one of the major application tasks.
\citet{li-etal-2021-compositional} examined the extent to which Transformer could correctly translate sentences with unseen combinations of known words.
\citet{dankers-etal-2022-paradox} evaluated whether models systematically translate sentences where some words were replaced with others and ones which were the combinations of sentences in the training set.
In addition to evaluations on synthetic datasets built with templates, they also conducted evaluations using a corpus to reflect the variations of vocabulary and syntactic structures in natural language.
\citet{moisio-etal-2023-using} split a corpus into the training and generalization sets using a distribution-based method proposed by \citet{keysers2020measuring}.
Their method distributes lexical items similarly and combinations of lexical items divergently between the training and generalization sets to create gaps for compositional generalization.

However, these studies focused mostly on lexical generalization, not structural generalization.
Furthermore, since evaluating structural generalization requires strict control over which lexical items and syntactic structures are seen/unseen during training, a distribution-based method is inappropriate here.
Therefore, we focus on both lexical and structural generalization and adopt a rule-based method to construct \dataset{}.

\section{Method}
\label{sec:method}
\subsection{Overview}
\label{sec:overview}
We create \dataset{} to evaluate the compositional generalization abilities of neural models in machine translation.
\dataset{} contains four sets of data: the training, development, test, and generalization sets.
The development and test sets are in-distribution sets, and the generalization set is an out-of-distribution set. 
The generalization set includes unseen lexical items and syntactic structures that are combinations of those in the training set.
This design ensures that models can translate these sentences correctly only when they succeed in translating compositionally according to lexical items and syntactic structures in the training set.

\subsection{Data Construction}
\label{sec:data-construction}
\begin{figure}[t]
    \centering
    \includegraphics[width=0.9\linewidth]{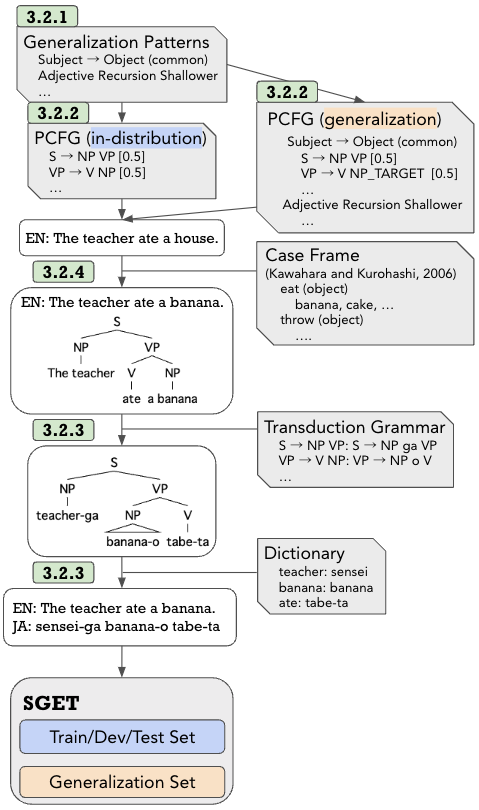}
    \caption{Data construction method.}
    \label{fig:construction}
\end{figure}

Figure~\ref{fig:construction} shows the data construction process, which consists of defining generalization patterns (Section~\ref{subsec:generalization-pattern}), generating English sentences using PCFGs (Section~\ref{subsec:pcfg}), creating English-Japanese parallel translation data using a rule-based method (Section~\ref{subsec:translation-data}), and filtering sentences (Section~\ref{subsec:selectional-restriction}).
This pipeline is also applicable to dataset generation for other language pairs by modifying the procedures correspondingly.

\subsubsection{Generalization Patterns}
\label{subsec:generalization-pattern}
\begin{table*}[t]
    \centering
    \small
    \begin{tabular}{cll}
    \toprule
    Category & Training & Generalization \\
    \midrule
    \multicolumn{3}{c}{Lexical Generalization}\\
    \midrule
    Primitive Substitution & The \textbf{goat} ate the apple. & The dog found the \textbf{goat}.\\\midrule
    Primitive & The man \textbf{moved} the cat. & \multirow{2}{*}{The tool \textbf{was moved}.}\\
    Structural Alternation & The plate \textbf{was used}.&\\\midrule
    Tense Alternation & The boy \textbf{offered} the girl the game. & The boy \textbf{offers} the girl the game.\\
    \midrule
    \multicolumn{3}{c}{Structural Generalization}\\
    \midrule
    Phrase  & The kid broke \textbf{the cup on the table}. & \multirow{2}{*}{\textbf{The baby in the room} cried.}\\
    Recombination& The child slept.&\\\midrule
    Recursion & James found the \textbf{small} cat. & \multirow{3}{*}{Noah loved the \textbf{unique big square new red} bag.}\\
    Depth Alternation & Ava bought a \textbf{rare blue} table. &\\
    &Lucas broke the \textbf{rare small red plastic} cup.&\\\midrule
    Gap Position & Liam knew \textbf{a kid} that \rule{0.3cm}{0.15mm} gave Ava a pen. & \multirow{2}{*}{Sam helped \textbf{the guy} that Ben gave a book to \rule{0.3cm}{0.15mm}.}\\
    Recombination & The chef knew \textbf{the guy} that Emily liked \rule{0.3cm}{0.15mm}. &\\\midrule
    \textit{wh}-question & \textbf{What} did the director buy? & \multirow{2}{*}{\textbf{What was thrown?}}\\
    Structural Alternation & The plate \textbf{was used}. &\\
    \bottomrule
    \end{tabular}
    \caption{Seven categories of generalization patterns.
    Each English sentence in the table is paired with its Japanese translation.}
    \label{table:pattern}
\end{table*}

We design generalization patterns for machine translation based on those for semantic parsing in COGS and SLOG.
We add 10 new patterns to the 32 existing patterns in COGS and SLOG to assess the models' generalization abilities in more detail.\footnote{We do not adopt 3 patterns in COGS that are difficult to evaluate in machine translation.}
Each pattern is defined so that the generalization sentences differ from those in the training set in specific aspects.
This allows us to evaluate how models can generalize to the unseen aspects.

We define seven categories of our generalization patterns, as given in Table~\ref{table:pattern}.
We describe the newly added categories and those mentioned in Section~\ref{sec:experiments}, and we explain the others in Appendix~\ref{sec:pattern_other}.

\paragraph{Tense Alternation}
English learners can understand and translate familiar verbs with a different tense form once they learn the basic rules of tense.
While existing benchmarks do not evaluate the generalization capacity of models to handle tense, it is an essential aspect of machine translation.
Therefore, we propose a generalization category for tense.

This category contains two patterns.
One is the generalization simply from the past tense form to the present tense form.
In this pattern, models should produce the present tense form of the verbs in the target language based on their suffixes.
The other is the generalization to the present tense form of target verbs in a certain argument structure when the training set contains their present tense form in a different argument structure and their past tense form in both argument structures.
This requires models to understand the relationship between an argument structure and a verb tense.
The difference between the two patterns is whether the translations of the verbs in the present tense are already given.
Both patterns require morphological generalization (i.e., generalization at the morphological level) in addition to lexical and structural generalization.

\paragraph{Phrase Recombination}
If a model has already learned the translation of a modifier in one grammatical role and can generalize compositionally, then it should be able to translate modifiers in another grammatical role correctly.
We test the generalization to those modifiers in the indirect object position or subject position, whereas the sentences in the training set contain them in the direct object position.
The modifiers that we evaluate are prepositional phrases (PPs), relative clauses (RCs), and adjectives.
COGS and SLOG performed evaluations on PPs and RCs, and both modifiers in an unseen position were challenging for the models.
Adjectives are shorter than PPs and RCs, and the word order of attributive adjective phrases is preserved in translation from English to Japanese.
Therefore, we add adjectives to assess whether they are easier to handle for the models.

We also topicalize 10\% of sentences with a modified phrase and include them in the training set, following \citet{wu-etal-2023-recogs}.
This enables models to learn that a modified phrase can appear in the beginning of a sentence.

\paragraph{Gap Position Recombination}
We evaluate the generalization to indirect object-extracted RCs based on subject and direct object-extracted ones in the training set.
We also test the generalization to a novel gap position of \textit{wh}-questions similarly.
\paragraph{Recursion Depth Alternation}
Recursions are essential in natural language because they make it possible to combine phrases to form more complex ones.
In theory, the number of recursions can be arbitrary finite, and they can form long and novel sentences, which plays a major role in linguistic competence~\citep{hauser2002faculty}.

We assess the generalization capacity of a model to translate sentences with an unseen recursion depth.
The recursion depth is the number of times that a syntactic structure is nested within the same structure.
We test four types of recursion: complement clauses (CPs), PPs, center-embedding of RCs, and adjectives.
We newly consider adjective patterns for the reason mentioned in the Phrase Recombination paragraph.
The training set includes recursions of depth 1, 2, and 4 in the training set and depth 3, 5, and 6 in the generalization set to allow assessments of generalization to shallower (1,2,4$\rightarrow$ 3) and deeper (1,2,4$\rightarrow$ 5,6).

We also concatenate sentences to make them longer than any sentence in the generalization set and include them in the training set, following \citet{wu-etal-2023-recogs}.
This eliminates the need for length generalization (i.e., generalization to longer sentences).

\subsubsection{Sentence Generation with PCFG}
\label{subsec:pcfg}

We generate English sentences using PCFGs, following COGS and SLOG, and we extend their PCFGs to support newly added patterns.
This rule-based method generates sentences with only the defined vocabulary and production rules, which allows us to control the generated sentences and the gaps between the training and generalization sets.
The PCFGs are detailed in Appendix~\ref{sec:detail_pcfg}.
To build an entire dataset, we generate sentences using the PCFG defined for the in-distribution sets and then split them into the training, development, and test sets.
For the generalization set, we generate sentences for each generalization pattern using an individually defined PCFG.

Also, we include primitive exposures (i.e., sentences with lexical items and syntactic structures not included in the training set but required for generalization) in the training set.

\subsubsection{Creating Translation Data}
\label{subsec:translation-data}

We translate English sentences generated with PCFGs into Japanese using RBMT, and for the latter we adopt the method proposed by \citet{wang-hershcovich-2023-evaluating}.
This allows us to control lexical items and syntactic structures in Japanese translations.
The translation process is shown in Figure~\ref{fig:construction}.
A sentence in the source language is parsed into a tree according to the production rules and then translated into a parse tree in the target language according to the transduction rules.
Finally, a translated sentence in the target language is obtained by converting the parsing tree into a sentence with the dictionary.

We use the production rules defined in Section~\ref{subsec:pcfg} as the ones in English for RBMT.
We manually create a dictionary and transduction rules between English and Japanese.
Note that we generally consider the basic word order (subject-object-verb) as the only correct translation.
We avoided considering flexibility of word order in Japanese~\citep{saito1985some}, which would unduly complicate the task and evaluation.
In addition, the transduction rules assign uniform translations to ambiguous sentences, following the previous studies on semantic parsing~\citep{kim-linzen-2020-cogs, li-etal-2023-slog}.
For instance, sentences such as \textit{The man noticed the girl in the house} with prepositional phrase attachment ambiguity are translated by considering the prepositional phrases as being attached to the noun phrase.

\subsubsection{Sentence Filtering}
\label{subsec:selectional-restriction}

Previous work has claimed that it is important to use naturally-occurring data that reflect variations in natural language in order to evaluate the compositional generalization abilities of neural models~\citep{shaw-etal-2021-compositional, dankers-etal-2022-paradox,moisio-etal-2023-using}.
However, sentences generated with PCFGs can be unnatural because, except for animacy constraints, we do not consider the relationships between lexical items in PCFGs; therefore, we either filter out those unnatural sentences or convert them into more natural ones.

First, we exclude sentences that have the same lexical item multiple times, such as \textit{The teacher liked the teacher}.
Then, we convert unnatural sentences into more natural ones by considering selectional restrictions (i.e., semantic constraints on combinations of lexical items).
We create a list by extracting pairs of a verb and a noun in Japanese that satisfy selectional restrictions from a Japanese case frame dictionary~\citep{kawahara-kurohashi-2006-case}.
We automatically check selectional restrictions for pairs between an inanimate subject and a verb and those between a verb and a direct object, utilizing the list.
If a sentence includes a pair that does not satisfy selectional restrictions, then we replace its noun with that in our list.
Figure~\ref{fig:construction} shows an example of such a replacement.

\section{Experiments}
\label{sec:experiments}
\subsection{Settings}
\label{sec:settings}
\subsubsection{Models}
\label{subsec:models}
We evaluate LSTM~\citep{hochreiter1997long}, vanilla Transformer~\citep{vaswani-2017-transformer}, and Llama~2~\citep{touvron2023llama}, with LSTM and Transformer trained from scratch using OpenNMT-py\footnote{\url{https://github.com/OpenNMT/OpenNMT-py}}~\citep{klein-etal-2017-opennmt}.
The reason why we focus on these three models is that evaluating them allows us to perform detailed and general analyses of their capabilities, and that the results can be transferred to most current models. All models have common features shared with various models, as vanilla Transformer is the core architecture of widely used state-of-the-art models, and Llama 2 is one of the standard open-source large language models.
In the preprocessing, Sudachi~\citep{TAKAOKA18.8884} is used to split Japanese sentences into words and Byte-Pair Encoding (BPE; ~\citealp{sennrich-etal-2016-neural}) is applied to both English and Japanese sentences; the number of subwords is set to 300 for Japanese and 650 for English.
To evaluate the models accurately, subword tokenization is necessary because without it, the target words of morphological generalization required in some generalization patterns are out-of-vocabulary.

We adopt relative positional encoding and disable early stopping and label smoothing because \citet{csordas-etal-2021-devil} showed that those settings benefit model performance in generalization.
Appendix~\ref{sec:detail_experiment} details the training settings, including the hyperparameters.

We train Transformer and Llama~2 five times each with randomly chosen seeds, and we use the average score of the five results as the final result to reduce the effect of randomness.
As for the model selection for evaluation, we choose the checkpoint with the best exact match accuracy on the development set because the accuracy on the generalization set is not always correlated with the validation loss~\citep{csordas-etal-2021-devil}.

\subsubsection{Dataset}
There are 43,800 sentences in the training set and 5,000 in each of the development and test sets.
The generalization set contains 76,000 sentences with 2,000 sentences per generalization pattern (except those related to ``CP recursion'' and wh-questions; see Appendix~\ref{sec:detail_pcfg} for details).
The number of primitive exposures is 100 for each generalization pattern, although COGS and SLOG use only one primitive exposure for each. 
We use 100 to ensure that the models learn the lexical items and syntactic structures in the primitive exposures, for which a single primitive exposure may be insufficient.
\subsubsection{Metrics}
\label{subsec:metrics}
We use three metrics to evaluate the models' translation results: Exact Match, BLEU, and Partial Match.
\paragraph{Exact Match}
Exact Match is defined as the proportion of sentences for which the model translation is exactly the same as the reference sentence generated using RBMT.
Using Exact Match to evaluate machine translation is usually inappropriate because multiple translations can be correct simultaneously.
However, Exact Match is a suitable metric for evaluating the overall performance on compositional generalization in this experiment for two reasons.
First, we design \dataset{} so that the correct translation of a sentence is uniquely determined.
Second, Exact Match considers the word order, which BLEU ignores.

\paragraph{BLEU}
We calculate BLEU scores using SacreBLEU~\citep{post-2018-call}.
Like Exact Match, BLEU is not a complete metric because it considers only the surface form of the words.
However, we use BLEU to evaluate the overall quality of translated sentences at the surface level.

\paragraph{Partial Match}
When a model incorrectly translates some words not directly related to a generalization pattern, the scores of Exact Match and BLEU become lower, which hinders direct evaluation of the generalization ability regarding the pattern.
Therefore, we propose Partial Match, which addresses only the constituents directly related to a specific generalization pattern.
We use Partial Match for generalization patterns whose target constituents can be determined automatically.
Note that we do not use it in some patterns (e.g., center embedding recursion) where evaluation of whole output sentences is appropriate.
It compares the target constituents in the translation sentence with the corresponding ones in the reference sentence, and it checks whether the target constituents have correct grammatical roles when the generalization pattern involves unseen combinations of grammatical roles and constituents.
Appendix~\ref{sec:detail_partial} details evaluation using Partial Match.

\subsection{Results}
\label{sec:results}

\begin{table}
    \centering
        
        \small
        \begin{tabular}{llcc}
            \toprule
            Model & Dataset & Exact (\%) & BLEU\\
            \midrule
            LSTM & Test & $93.4_{(\pm 3.4)}$ & $96.2_{(\pm 2.2)}$\\
            & Gen. & $9.6_{(\pm 4.8)}$ & $64.9_{(\pm 8.3)}$\\
            \midrule
            Transformer & Test & $99.3_{(\pm 0.1)}$ & $99.6_{(\pm 0.0)}$\\
            & Gen. & $48.7_{(\pm 1.6)}$ & $84.0_{(\pm 0.6)}$\\
            \midrule
            Llama~2 & Test & $99.4_{(\pm 0.1)}$ & $99.8_{(\pm 0.1)}$\\
            & Gen. & $80.6_{(\pm 0.5)}$ & $95.5_{(\pm 0.2)}$\\
            \bottomrule
        \end{tabular}
        \caption[Results on test and generalization set.]{Results on test and generalization sets.}
        \label{table:overall_result}
\end{table}

\subsubsection{Overall Results}

Table~\ref{table:overall_result} gives the overall results on the test and generalization sets.
Both Transformer and Llama~2 achieved near-perfect scores on the test set, which confirms that they learned the lexical items and syntactic structures in the training set well.
On the other hand, the scores of Transformer on the generalization set, especially Exact Match, are about 50\% lower than those on the test set.
This shows that Transformer struggles with compositional generalization.
Llama~2 outperformed Transformer on the generalization set, but a gap remains between the test and generalization scores.

However, we cannot necessarily conclude that Llama~2 has better compositional generalization capacity than LSTM and Transformer based on the results on the generalization set.
\citet{kim2022uncontrolled} claimed that the compositional generalization capacity of pretrained models is overestimated because in pretraining they may learn lexical items and syntactic structures that are subsequently questioned in the generalization set.
Therefore, the present results might be because Llama~2 is a pretrained model whereas Transformer is trained from scratch.

In fact, it is difficult to analyze rigorously how pretraining data impact the structural generalization capacity of models.
One way to do so would be to use sentences whose syntactic structure is definitely unseen for a pretrained model, but it is challenging to control the distribution of seen and unseen syntactic structures in the pretraining data for large language models.

\begin{table}[t]
    \centering
        
        \small
        \begin{tabular}{llcc}
            \toprule
            Model & Group & Exact (\%) & BLEU\\
            \midrule
            Transformer & Lexical & $58.2_{(\pm 1.7)}$& $82.7_{(\pm 0.9)}$\\
            & Lex.\ + Mor. & $40.5_{(\pm 2.7)}$& $80.6_{(\pm 0.9)}$\\
            & Structural & $45.2_{(\pm 2.3)}$ & $83.3_{(\pm 0.5)}$\\
            \midrule
            Llama~2 & Lexical & $87.0_{(\pm 0.7)}$ & $96.8_{(\pm 0.2)}$ \\
            & Lex.\ + Mor. & $72.7_{(\pm 1.1)}$ & $91.8_{(\pm 0.4)}$\\
            & Structural & $79.4_{(\pm 0.9)}$ & $94.3_{(\pm 0.5)}$\\
            \bottomrule
        \end{tabular}
        \caption[Result by generalization groups.]{Results by generalization groups.}
        \label{table:group_result}
\end{table}

\begin{table}
    \centering
        
        \small
        \begin{tabular}{lcccc}
            \toprule
             Pattern & Transformer & Llama~2\\
            \midrule
            PP in Subj & $11.9_{(\pm 6.2)}$ & $93.6_{(\pm 2.3)}$ \\
            PP in indirect Obj & $18.9_{(\pm 7.1)}$ & $92.5_{(\pm 2.7)}$\\
            RC in Subj & $1.2_{(\pm 0.7)}$ & $75.9_{(\pm 5.2)}$\\
            RC in indirect Obj & $3.7_{(\pm 0.8)}$ & $54.9_{(\pm 1.9)}$\\
            Adj in Subj & $68.4_{(\pm 7.6)}$ & $96.5_{(\pm 0.8)}$\\
            Adj in indirect Obj & $48.8_{(\pm 7.4)}$ & $98.8_{(\pm 0.4)}$\\
            \bottomrule
        \end{tabular}
        \caption{Results of Exact Match (\%) in phrase recombination.}
        \label{table:result_phrase_recursion}
\end{table}

\subsubsection{Results by Generalization Pattern}
\label{subsec:results-by-pattern}

We focus on Transformer-based models from here, as LSTM performed extremely poorly.
First, we analyze the tendency of the scores across all the patterns.
We classify the patterns into three groups: lexical generalization with morphological generalization, lexical generalization without morphological generalization, and structural generalization.
The scores regarding these three groups are given in Table~\ref{table:group_result}.
Transformer and Llama~2 scored highest in the lexical generalization without morphological generalization and performed worse in structural generalization.

This tendency is consistent with the results of previous studies~\citep{kim-linzen-2020-cogs, li-etal-2023-slog}, although the difference between the two groups is smaller in our results.
There are two possible reasons for this.
One is that machine translation differs from semantic parsing in output formats, and the other is that we updated the experimental settings for fairer evaluation of structural generalizations.
\paragraph{Phrase Recombination}
Table~\ref{table:result_phrase_recursion} gives the results in this category.
Transformer struggled in generalizations to PPs in unseen grammatical roles, whereas it performed better in translating adjectives in unseen grammatical roles.
This suggests that the word order of a modifier and a noun between English and Japanese influences the generalization performance of this category: the word order is maintained in translating adjectives, whereas a PP is placed after the modified noun in English but before it in Japanese.
\paragraph{Recursion Depth Alternation}
Table~\ref{table:length_result} gives the results in this category.
Transformer struggled in generalization to a novel recursion depth except in adjective recursions, in which it achieved near-perfect scores.
We argue that generalization to adjective recursions is easier because both the word order and the number of words are maintained in translation, as mentioned in Section~\ref{subsec:generalization-pattern}.
This result suggests that Transformer does not struggle in every recursion type.

\subsection{Discussion}
\label{sec:discussion}

\subsubsection{Difference Between Machine Translation and Semantic Parsing}
\label{sec:task}
\dataset{} is focused on machine translation, whereas existing datasets such as COGS and SLOG are focused on semantic parsing.
This difference in task settings seems to have resulted in differing model performance in some generalization patterns.

One such case is the generalization to indirect object-extracted RCs.
In this pattern, Transformer scored 43.2\% in Exact Match in \dataset{}, while it scored only 4.7\% in SLOG.
In semantic parsing, Transformer handled indirect object-extracted RCs as direct object-extracted ones in most cases, while it handled them as subject-extracted ones in errors in machine translation, and those errors are fewer.
\citet{li-etal-2023-slog} attributed the failure of Transformer in this pattern in semantic parsing to the decoder being prone to generating often-seen constituents.
However, the translation of an indirect object-extracted RC does not appear in the training set, yet the model was able to generate it.
One possible reason is that Transformer has to generate a novel semantic representation in semantic parsing, whereas the subsequences of a correct translation are in the training set in machine translation, which makes the generalization easier.

\begin{table}
    \centering
        
        \small
        \begin{tabular}{llcc}
            \toprule
            Restrictions & Dataset & Exact (\%) & BLEU\\
            \midrule
            Yes & Test & $99.4_{(\pm 0.1)}$ & $99.8_{(\pm 0.1)}$\\
            & Gen. & $80.6_{(\pm 0.5)}$ & $95.5_{(\pm 0.2)}$\\
            \midrule
            No & Test & $98.9_{(\pm 0.4)}$ & $99.7_{(\pm 0.1)}$\\
            & Gen.& $79.1_{(\pm 0.3)}$ & $95.1_{(\pm 0.1)}$\\
            \bottomrule
        \end{tabular}
        \caption[Comparison of results of Llama~2 fine-tuned with dataset with/without selectional restrictions.]{Comparison of results of Llama~2 fine-tuned with dataset with/without selectional restrictions.}
        \label{table:restriction_result}
\end{table}

\subsubsection{Effects of Selectional Restriction}
\label{sec:selectional}

We introduced selectional restrictions into our dataset to avoid generating unnatural sentences, as mentioned in Section~\ref{subsec:selectional-restriction}.
However, it is unclear to what extent the model performance in generalization is affected by the naturalness of the sentences.
Therefore, we evaluate Llama~2 on our dataset without selectional restrictions and compare the results to those with selectional restrictions.

Table~\ref{table:restriction_result} compares the results of Llama~2 fine-tuned with the dataset with and without selectional restrictions.
The results show that selectional restrictions had a negligible impact on the performance of Llama~2, and so pretraining seems not to have prevented the model from translating sentences that contained unnatural combinations of words.

This result suggests that a synthetic dataset is sufficiently good for evaluating compositional generalization abilities, although some studies underline the importance of evaluation with real-world data~\citep{dankers-etal-2022-paradox, moisio-etal-2023-using}.
Note that selectional restrictions are not applied thoroughly in our dataset, which might have affected these evaluation results.
Further investigation is needed to confirm the impact of selectional restrictions, which we leave for future work.

\begin{table*}[t]
    \centering
        \small
        \begin{tabular}{lcccc}
            \toprule
            & \multicolumn{2}{c}{Transformer} & \multicolumn{2}{c}{Llama~2}\\
            Generalization pattern & w.Concat & wo.Concat & w.Concat & wo.Concat\\
            \midrule
            CP recursion shallower &
            $79.8_{(\pm 2.7)}$ & $53.1_{(\pm 3.9)}$ &
            $97.4_{(\pm 0.6)}$ & $96.5_{(\pm 1.0)}$\\
            CP recursion deeper &
            $6.2_{(\pm 1.4)}$ & $0.7_{(\pm 0.3)}$ &
            $82.0_{(\pm 3.6)}$ & $59.2_{(\pm 6.7)}$\\
            PP recursion shallower &
            $71.5_{(\pm 2.8)}$ & $49.4_{(\pm 5.2)}$ &
            $95.8_{(\pm 0.6)}$ & $91.1_{(\pm 1.0)}$\\
            PP recursion deeper &
            $10.5_{(\pm 1.5)}$ & $2.4_{(\pm 1.4)}$ &
            $81.8_{(\pm 1.9)}$ & $59.2_{(\pm 2.5)}$\\
            CE recursion shallower &
            $73.3_{(\pm 5.9)}$ & $31.3_{(\pm 4.5)}$ &
            $96.0_{(\pm 1.4)}$ & $94.9_{(\pm 1.1)}$ \\
            CE recursion deeper &
            $11.7_{(\pm 3.2)}$ & $0.6_{(\pm 0.3)}$ &
            $75.9_{(\pm 4.0)}$ & $56.8_{(\pm 6.8)}$\\
            Adj recursion shallower &
            $99.2_{(\pm 0.3)}$ & $97.9_{(\pm 0.7)}$ &
            $98.7_{(\pm 0.4)}$ & $98.5_{(\pm 0.4)}$\\
            Adj recursion deeper &
            $99.5_{(\pm 0.1)}$ & $98.5_{(\pm 0.6)}$ &
            $98.4_{(\pm 0.5)}$ & $97.6_{(\pm 1.0)}$\\
            \bottomrule
        \end{tabular}
        \caption[Comparison of Exact Match in patterns regarding recursions between models trained with/without concatenated sentences.]{Comparison of Exact Match (\%) in patterns regarding recursions between models trained with/without concatenated sentences.
        ``w.Concat'' (resp.\ ``wo.Concat'') means that the model was trained with (resp.\ without) concatenated sentences.}
        \label{table:length_result}
\end{table*}

\begin{table}[t]
    \centering
        \small
        \begin{tabular}{llcccc}
            \toprule
            Length & Concat. & Transformer & Llama~2\\
            \midrule
            Longer & w.Concat & $4.1_{(\pm 0.6)}$ & $72.0_{(\pm 0.6)}$\\
            Longer & wo.Concat & $0.6_{(\pm 0.2)}$ & $42.4_{(\pm 1.0)}$\\
            Shorter & w.Concat & $6.7_{(\pm 1.6)}$ & $84.3_{(\pm 3.6)}$\\
            Shorter & wo.Concat & $0.7_{(\pm 0.3)}$ & $63.1_{(\pm 6.7)}$\\
            \bottomrule
        \end{tabular}
        \caption[Comparison of scores in ``CP recursion deeper" by length.]{Comparison of scores in ``CP recursion deeper" by length. ``Longer" (resp.\ ``Shorter'') means that the length of a source sentence is longer than or equal to (resp.\ shorter than) the maximum length of the training set.}
        \label{table:length_comp_result}
\end{table}

\subsubsection{Length Generalization}
\label{sec:length}

As mentioned in ``Recursion Depth Alternation'' in Section~\ref{subsec:generalization-pattern}, we added concatenated sentences to the training set to ensure that length generalization is not required for the models.
We investigate whether doing so improves the performance of the models in recursion generalizations in our dataset by comparing the results of the models trained with and without concatenated sentences.

Table~\ref{table:length_result} gives the results of Transformer and Llama~2 trained or fine-tuned with/without concatenated sentences in recursion generalizations.
The models trained without concatenated sentences performed worse than those trained with concatenated sentences in all recursion patterns, and a max of 40\% dropoff in Exact Match was seen.
Table~\ref{table:length_comp_result} compares how the model performance in ``CP recursion deeper'' varied depending on the sentence length.
It shows that the models trained without concatenation performed worse than those trained with concatenation, not only in longer sentences but also in shorter ones.
These results suggest that adding concatenated sentences enhanced the models' ability to generalize to long sentences in general, thereby improving their performance in recursion patterns.

\section{Conclusion}
\label{sec:conclusion}

In this paper, we constructed \dataset{} for evaluating the compositional generalization abilities of neural models in English-Japanese machine translation.
We generated the dataset via a rule-based approach, which allowed us to control the lexical items and syntactic structures in the dataset and strictly evaluate the compositional generalization abilities of models.
We then used the generated dataset to investigate the compositional generalization abilities of LSTM, vanilla Transformer, and Llama~2.

We found that all models struggled to generalize compositionally to unseen constituents overall, while they achieved near-perfect scores on the in-distribution test set.
The results also suggested that for the models, structural generalization is more challenging than lexical generalization.
These findings regarding machine translation are consistent with the results of previous studies on semantic parsing.
However, the models correctly filled the structural gap questioned in some generalization patterns in which the models performed poorly in semantic parsing, which might have been because of the fairer settings that we adopted and the difference between the two tasks.
This underlines the importance of evaluating the compositional generalization abilities of models under diverse settings, not only in semantic parsing but also in other tasks such as machine translation.
We also discovered that the naturalness of the sentences in \dataset{} had little influence on the performance of Llama~2.
This indicates that a synthetically generated benchmark is adequate for evaluating compositional generalization in a controlled manner.

Future work might focus on comprehensive evaluations across various languages and tasks.
The RBMT method adopted in our approach originally handles English-Chinese translation in addition to English-Japanese~\citep{wang-hershcovich-2023-evaluating}, so targeting other languages, including Chinese, should be feasible.

\section*{Limitations}
\label{sec:limitations}

\subsection*{Selectional Restrictions}
We introduced selectional restrictions to \dataset{}, but there are two limitations because of our data construction method.
First, we do not consider pairs other than those between a verb and a noun (e.g., those between a prepositional phrase and a verb) because the case frame dictionary does not contain them.
Second, in some cases a noun is involved in multiple pairs, and we cannot find a noun satisfying all selectional restrictions.
We handle such cases by replacing the noun with another one that satisfies selectional restrictions for one of the pairs.
For example, below is a sentence in which \textit{bed} is involved with two pairs, \textit{bed} and \textit{eat}, and \textit{bed} and \textit{tell}.
\begin{exe}
    \ex The teacher ate the bed that the teacher told to me.
\end{exe}
In this case, we cannot find a noun that satisfies selectional restrictions for both pairs.
We settle on replacing \textit{bed} with \textit{apple}, which does not violate selectional restrictions with respect to \textit{apple}.
These limitations force us to leave some unnatural sentences in the dataset, and we leave the solution for future work.

\subsection*{Task Settings}
As discussed in Section~\ref{sec:task}, the translation task settings can impact the model performance.
We adopt English-Japanese translation as the target task in our dataset for two reasons.
The first is that English and Japanese have a different surface order of words even when the structure is the same, so translation between the two languages requires understanding the structure.
The second is that Japanese has postpositional particles, which show the grammatical role of the previous word and are suitable for analyzing compositional generalization.
However, translation between other pairs of languages may reveal other aspects of the models' compositional generalization abilities, which we leave for future work.

\subsection*{Experimental Settings}
We did not tune the number of subwords for BPE.
This can impact model performance in generalization because the size of the subword vocabulary determines how input words are split.
There have been few studies on the impact of BPE on compositional generalization, and further experiments are needed on this topic, which we also leave for future work.

\section*{Acknowledgements}
We thank the three anonymous reviewers and the meta-reviewer for providing helpful comments and feedback.
This work was supported by PRESTO, JST Grant Number JPMJPR21C8, Japan.

\bibliography{anthology,custom}

\appendix

\begin{table*}[t]
    \centering
    
    \footnotesize
    \begin{tabular}{p{0.305\linewidth}p{0.295\linewidth}p{0.32\linewidth}}
    \toprule
    Pattern & Training & Generalization \\
    \midrule
    \multicolumn{3}{c}{Primitive Substitution}\\
    \midrule
    Subj $\rightarrow$ Obj (common)& 
     The \textbf{goat} ate the apple. &
     The dog found the \textbf{goat}. \\
     Subj $\rightarrow$ Obj (proper)&
     \textbf{Chris} liked the woman. &
     The baby gave \textbf{Chris} the key.\\
     Obj $\rightarrow$ Subj (common) &
     The woman found the \textbf{panda}. & 
     The \textbf{panda} ran.\\
     Obj $\rightarrow$ Subj (proper) &
     The man liked \textbf{Taylor}. &
     \textbf{Taylor} bought Oliver the bag.\\
     Prim $\rightarrow$ Subj (common) &
     \textbf{thief}&
     The \textbf{thief} fell.\\
     Prim $\rightarrow$ Subj (proper) &
     \textbf{Coco}&
     \textbf{Coco} ate.\\
     Prim $\rightarrow$ Obj (common) &
     \textbf{trainer}&
     The friend found the \textbf{trainer}.\\
     Prim $\rightarrow$ Obj (proper) &
     \textbf{Nova}&
     A book was passed to \textbf{Nova}.\\
     Prim $\rightarrow$ Infinitive (verb) &
     \textbf{jump} &
     Mason decided to \textbf{jump}.\\
     \midrule
     \multicolumn{3}{c}{Tense Alternation}\\
     \midrule
    Past$\rightarrow$ Present (Ditransitive) &
    Ava \textbf{lent} the doll to Lucas. &
    Olivia \textbf{lends} a box to the kid.\\
    Past$\rightarrow$ Present (Infinitive) &
    A child \textbf{wanted} to run. &
    A child \textbf{wants} to talk.\\
    Past$\rightarrow$ Present (Complement) &
    The host \textbf{hoped} that Noah grew. &
    A hero \textbf{hopes} that Olivia was helped.\\
    \midrule
    Past, Present (Transitive) &&\\
    $\rightarrow$ Present (Ditransitive) &
    The friend \textbf{showed} a guitar. / Noah \textbf{showed} the book to the mother. / Sam \textbf{shows} a big bowl.&
    A patient \textbf{shows} the father a cup.\\
    $\rightarrow$ Present (Infinitive) &
    The child \textbf{prepared} a cup. / A boy \textbf{prepared} to walk. / The guest \textbf{prepares} the box.&
    The mother \textbf{prepares} to talk.\\
    $\rightarrow$ Present (Complement) &
    A child \textbf{learned} the newspaper. / Emma \textbf{learned} that a shoe broke. / A student \textbf{learns} a book.&
    A girl \textbf{learns} that the cup was found.\\
    \midrule
    \multicolumn{3}{c}{Primitive Structural Alternation}\\
    \midrule
     Active $\rightarrow$ Passive &
    The man \textbf{moved} the car.&
    The tool was \textbf{moved}.\\
    Passive $\rightarrow$ Active &
    The apple was \textbf{dropped}.&
    The child \textbf{dropped} the book.\\
    Object-omitted Transitive $\rightarrow$ Transitive &
    John \textbf{wrote}.&
    The bunny \textbf{wrote} the book.\\
    Unaccusative $\rightarrow$ Transitive &
    The apple \textbf{exploded}.&
    Lucy \textbf{exploded} the ball.\\
    Double Object $\rightarrow$ PP &
    John \textbf{granted} Mary the book.	&
    The teacher \textbf{granted} the plate to the child.\\
    PP $\rightarrow$ Double Object &
    John \textbf{gifted} the book to Mary.&
    Ava \textbf{gifted} the friend the chair.\\
    \bottomrule
    \end{tabular}
    \caption[Lexical generalization patterns.]{Lexical generalization patterns.}
    \label{table:pattern_lexical}
\end{table*}

\begin{table*}[t]
    \centering
    
    \footnotesize
    \begin{tabular}{p{0.302\linewidth}p{0.305\linewidth}p{0.315\linewidth}}
    \toprule
    Pattern & Training & Generalization \\
    \midrule
    \multicolumn{3}{c}{Phrase Recombination}\\
    \midrule
    PP in Direct Obj $\rightarrow$ Subj&
     John gave \textbf{the pen on the seat} to Ava. &
     \textbf{The book beside the bed} fell.\\ 
     PP in Direct Obj$\rightarrow$ Indirect Obj&
     John gave \textbf{the pen on the seat} to Ava.&
     Noah gave a bag to \textbf{the kid in the house}.\\
     RC in Direct Obj$\rightarrow$ Subj &
     The dog gave Liam \textbf{the apple that John liked}.&
     \textbf{The kid that found Emma} ate an apple.\\
     RC in Direct Obj$\rightarrow$ Indirect Obj &
     The dog gave Liam \textbf{the apple that John liked}.&
     The dog gave the apple to the kid that Liam loved.\\
     Adj in Direct Obj$\rightarrow$Subj&
     The driver found \textbf{the tall girl}.&
     \textbf{The tall girl} found the driver.\\
     Adj in Direct Obj$\rightarrow$Indirect Obj &
     The driver found \textbf{the tall girl}. &
     The doctor gave the book to \textbf{the tall girl}.\\
     \midrule
    \multicolumn{3}{c}{Recursion Depth Alternation}\\
    \midrule
    CP recursion shallower: \newline depth 1, 2, 4 $\rightarrow$ depth 3&
     The kid admired \textbf{that} Liam dreamed \textbf{that} Sam was helped by the horse.
     & Samuel believed \textbf{that} Liam thought \textbf{that} the men knew \textbf{that} Ava packed the fig.\\
     CP recursion deeper: \newline depth 1, 2, 4 $\rightarrow$ depth 5, 6 &
     The kid admired \textbf{that} Liam dreamed \textbf{that} the friend was helped by the horse. &
     A patient thought \textbf{that} Liam said \textbf{that} a tenant meant \textbf{that} a girl proved \textbf{that} Olivia wished \textbf{that} Ava ran.\\
     PP recursion shallower: \newline depth 1, 2, 4 $\rightarrow$ depth 3 &
     A soldier found the bottle \textbf{beside the book on the table}. &
     Noah found a book \textbf{on the table in the room in the house}.\\
     PP recursion deeper: \newline depth 1, 2, 4 $\rightarrow$ depth 5, 6&
     A soldier found the bottle \textbf{beside the book on the table}. &
     Noah found a book \textbf{beside the cup on the table in the room in the house on the road}.\\
     CE recursion shallower: \newline depth 1, 2, 4 $\rightarrow$ depth 3 &
     A bird sought a book \textbf{that} the child \textbf{that} a mother knew found. &
     Liam found a cup \textbf{that} a monkey \textbf{that} a guy \textbf{that} Ava knew observed broke.\\
     CE recursion deeper: \newline depth 1, 2, 4 $\rightarrow$ depth 5, 6 &
     A bird sought a book \textbf{that} the child \textbf{that} a mother knew found. &
     Ava loved a cat \textbf{that} the guy \textbf{that} the fish \textbf{that} the visitor \textbf{that} the child \textbf{that} Olivia called discovered drew heard knew.\\	
     Adj recursion shallower: \newline depth 1, 2, 4 $\rightarrow$ depth 3 &
     Ava bought a \textbf{rare blue} table.&
     James called the \textbf{calm old English} friend.\\
     Adj recursion deeper: \newline depth 1, 2, 4 $\rightarrow$ depth 5, 6 &
     Ava bought a \textbf{rare blue} table. &
     Noah noticed \textbf{the beautiful big square new red} bag.\\
     \midrule
    \multicolumn{3}{c}{Gap Position Recombination}\\
    \midrule
    Subj, Direct Obj-extracted RC \newline $\rightarrow$ Indirect Obj-extracted RC &
    Ava knew \textbf{the guy that Liam liked}. / 
    Liam knew \textbf{a kid that gave the guest a pen}. &
    The visitor liked \textbf{the guy that the friend gave the plant to}.\\
    Subj, Direct Obj-extracted \textit{wh}-question \newline $\rightarrow$ Indirect Obj-extracted \textit{wh}-question &
    \textbf{Who appreciated} the book? /
    \textbf{Who} did Liam \textbf{call}? &
    \textbf{Who} did Liam \textbf{give} a shoe \textbf{to}?\\
    \midrule
     \multicolumn{3}{c}{\textit{Wh}-question Structural Alternation}\\
     \midrule
     Transitive&&\\
    $\rightarrow$ Active Subject &
    \textbf{Who broke} the cup? &
    \textbf{Who wanted} to talk?\\
    $\rightarrow$ Passive Subject &
    \textbf{Who broke} the cup?&
    \textbf{What was broken}?\\
    $\rightarrow$ Direct Obj, Ditransitive &
    \textbf{What} did the girl \textbf{break}?&
    \textbf{What} did the girl \textbf{give} to the boy?\\
    $\rightarrow$ Subject with PP &
    What did \textbf{the girl} break?&
    What did \textbf{the boy in the car} eat?\\
    $\rightarrow$ Long movement &
    \textbf{What} did the girl \textbf{break}?&
    \textbf{What} did Liam think that the girl \textbf{break}?\\
         \bottomrule
    \end{tabular}
    \caption[Structural generalization patterns.]{Structural generalization patterns.}
    \label{table:pattern_structural}
\end{table*}

\section{Other Generalization Patterns}
\label{sec:pattern_other}

The following categories are the same as those in COGS and SLOG, and the names of some categories are taken from \citet{an-etal-2023-context}.
Table~\ref{table:pattern_lexical} lists the lexical generalization patterns and Table~\ref{table:pattern_structural} lists the structural generalization patterns.

\subsection{Primitive Substitution}
If humans can translate a noun in a particular grammatical role, then they can translate the noun in a different one. 
Similarly, if humans can translate a noun itself, then they can translate the noun in a sentence.
In this category, we evaluate the generalization capacity of models to translate a familiar word in a grammatical role in which the word is unseen.

In the patterns that require generalization from one grammatical role to another, their target words appear only in a certain position in the in-distribution sets and in a different position in the generalization set.
Also, in the patterns that require generalization from a word alone to a word in a sentence, their target words appear alone in the in-distribution sets and in a certain position in the generalization set.

\subsection{Primitive Structural Alternation}
A verb can have different argument structures.
For example, \textit{eat} can function as a transitive or an intransitive verb and can also be used in the active or passive voice.
If humans are familiar with a verb in a particular argument structure and the basic rules of argument structures, then they can translate the verb in a different argument structure.
We test the models' generalization ability to translate verbs with an argument structure with which they are unseen.
We select target verbs from those whose past and past participle forms in English are the same in order not to require morphological generalizations.
The training set includes the target verbs with a particular argument structure only, and the generalization set includes ones with another argument structure according to the pattern. 
Also, the training set has argument structures that are questioned in the generalization set with other verbs.

\subsection{\textit{Wh}-question Structural Alternation}
Humans can understand unseen structures of \textit{wh}-questions based on similar ones in declarative sentences.
All the \textit{wh}-questions in the training set have the subject-verb-object (SVO) order as their main structure, and generalization sentences for each pattern have structures that are different from SVO.
Also, the training set contains structures in declarative sentences corresponding to unseen ones in \textit{wh}-questions.

\section{PCFG}
\label{sec:detail_pcfg}

Defining a PCFG involves selecting lexical items and defining production rules.
We select lexical items based on those in COGS and SLOG, excluding some grammatically incorrect word usages therein, such as \textit{enjoy to do} and \textit{like that}. 
We also add 43 adjectives to introduce new generalization patterns related to adjectives.
To increase the number of the target words for each generalization pattern, as explained in Section~\ref{subsec:generalization-pattern}, we add 32 nouns and 28 verbs.
The final vocabulary contains 123 proper nouns, 423 common nouns, 178 verbs, and 43 adjectives.

In defining each grammar, we ensure a desired gap between the training and generalization sets.
The training set should include sentences that are required for specific generalizations, not those that can be direct solutions to translating sentences in the generalization set without any generalization.
The probabilities assigned to the production rules in PCFGs are assigned similarly to COGS and SLOG, although they are adjusted to reflect the addition of new rules.
The probabilities assigned to the lexical items follow Zipf's law.

Also, we set target words (i.e., unseen words for evaluating generalization) for the patterns that require lexical generalization.
Although COGS assigns only one target word for each pattern, SGET assigns five target words for each to decrease the effect of the word choice.
In addition, for every pattern except ``CP Recursion'' in ``Novel Recursion Depth'' and patterns related to \textit{wh}-questions, we add sentences with the target constituent in complement clauses to the generalization set.
Patterns related to ``CP recursion'' and \textit{wh}-questions have 1,000 sentences each in the generalization set because they do not include sentences with the target constituent in complement clauses.
Translating these sentences is more challenging for models because sentences with the target constituent in a complement clause have more complicated structures than those without a complement clause.
As for the target verbs in ``Novel Tense'', the target verbs of each pattern consist of four regular verbs and one irregular verb.
Translating an irregular verb in an unseen tense form requires additional morphological generalization, so we control the target verbs to balance the effect of irregular verbs among the patterns.

\section{Training Details}
\label{sec:detail_experiment}
\paragraph{LSTM}
We use a four-layer encoder-decoder and adopt global attention and a dot-product score function.
We set the learning rate as 1e-4, the batch size as 256, and the number of training steps as 70,000.
\paragraph{Transformer}
We use Adam~\citep{adam-2015-kingma} as the optimizer and set the learning rate as 1e-4, the batch size as 256, and the number of training steps as 70,000.
Our Transformer model has six encoder and six decoder layers, and eight attention heads and was trained for four hours on a single GPU.
\paragraph{Llama~2}
We fine-tune Llama~2\footnote{\url{https://huggingface.co/meta-llama/Llama-2-7b-hf}} with LoRA~\citep{hu2022lora}. 
We set the learning rate as 1e-4, LoRA $r$ as 8, LoRA $\alpha$ as 32, the dropout rate as 0.1, the batch size as 64, and the number of epochs as 8.
We fine-tuned Llama~2 for 12 hours on a single GPU.

\section{Partial Match in Detail}
\label{sec:detail_partial}
For example, in the generalization pattern ``Subj $\rightarrow$ Obj (common)'' in Table~\ref{table:pattern_lexical}, consider evaluating Example~\ref{ex:partial} below.
One of the target words in this pattern is \textit{panda}, and Partial Match checks whether \textit{panda} is translated correctly and is in the object position.
\begin{exe}
    \ex\label{ex:partial}\begin{xlist}
    \exi{En:} The woman found the \textbf{panda}.
    \exi{Ja-Gold:} \gll jyosei-ga \textbf{panda}-o mituke-ta\\
    {woman-\Nom} {panda-\Acc} {find-\Pst}\\
    \exi{Ja-Pred:} \, \begin{xlist}
        \exi{(i)} \gll jyosei-ga \textbf{inu}-o mituke-ta\\
        {woman-\Nom} {dog-\Acc} {find-\Pst}\\
        \exi{(ii)} \gll panda-ga \textbf{jyosei}-o mituke-ta\\
        {panda-\Nom} {woman-\Acc} {find-\Pst}\\
        \exi{(iii)} \gll dansei-ga \textbf{panda}-o mituke-ta\\
        {man-\Nom} {panda-\Acc} {find-\Pst}\\
    \end{xlist}
    \end{xlist}
\end{exe}
Partial Match considers (i) and (ii) to be incorrect and (iii) to be correct.
(i) does not include the correct translation of the target word, and (ii) does not have the target word in the correct grammatical role.
(iii) is considered correct because it satisfies both requirements, although it is not the same as the reference sentence.

To analyze the phrase structure of a translation sentence and obtain grammatical roles therein, we use GiNZA\footnote{\url{https://github.com/megagonlabs/ginza}}, a phrase structure analyzer for Japanese.

\section{Results by Generalization Pattern in Detail}
\label{sec:result_other}
\begin{table*}
\centering
    \footnotesize
    \begin{tabular}{lcccccc}
    \toprule
    &
    \multicolumn{3}{c}{Transformer} &
    \multicolumn{3}{c}{Llama~2} \\
    Generalization Pattern &
    Exact & BLEU & Partial &
    Exact & BLEU & Partial\\
    \midrule
    \multicolumn{7}{c}{Primitive Substitution}\\
    \midrule
    Subj $\rightarrow$ Obj (common) &
    91.4 & 97.4 & 91.8 &
    98.7 & 99.7 & 99.2\\
    Subj $\rightarrow$ Obj (proper) &
    94.9 & 98.6 & 95.2 &
    98.3 & 99.5 & 99.0\\
    Obj $\rightarrow$ Subj (common) &
    95.5 & 98.8 & 98.7 &
    96.6 & 99.7 & 99.4 \\
    Obj $\rightarrow$ Subj (proper) &
    97.2 & 99.1 & 99.4 &
    98.7 & 99.4 & 99.9\\
    Prim $\rightarrow$ Subj (common) &
    12.8 & 35.2 & 13.0 &
    71.2 & 92.3 & 71.6\\
    Prim $\rightarrow$ Subj (proper) &
    12.8 & 68.1 & 13.0 &
    94.7 & 98.4 & 95.9\\
    Prim $\rightarrow$ Obj (common) &
    5.7 & 51.2 & 6.9 &
    71.2 & 92.3 & 71.6\\
    Prim $\rightarrow$ Obj (proper) &
    22.1 & 77.0 & 22.2 &
    69.9 & 92.0 & 70.4\\
    Prim $\rightarrow$ Verb &
    40.5 & 71.9 & --- &
    14.8 & 60.6 & ---\\
    \midrule
    \multicolumn{7}{c}{Primitive Structural Alternation}\\
    \midrule
    Active $\rightarrow$ Passive &
    63.6 & 81.5 & 63.8&
    78.3 & 90.5 & 78.7\\
    Passive $\rightarrow$ Active &
    48.6 & 83.7 & 66.9&
    94.3 & 98.2 & 99.9\\
    Obj-omitted transitive $\rightarrow$ Transitive &
    53.9 & 83.7 & 63.9&
    78.6 & 92.2 & 79.2\\
    Unaccusative $\rightarrow$ Transitive &
    45.4 & 82.5 & 46.6 &
    74.0 & 92.2 & 74.5\\
    Double Obj $\rightarrow$ PP &
    98.3 & 99.6 & 99.8&
    99.0 & 99.7 & 99.8\\
    PP $\rightarrow$ Double Obj &
    93.5 & 97.7 & 97.3 &
    96.4 & 99.3 & 97.0\\
    \midrule
    \multicolumn{7}{c}{Tense Alternation}\\
    \midrule
    Present (Ditransitive) &
    31.3 & 80.2 & 34.4&
    96.8 & 99.3 & 98.6\\
    Present (Transitive $\rightarrow$ Ditransitive) &
    69.2 & 90.5 & 85.7&
    96.4 & 99.2 & 98.7\\
    Present (Infinitive) &
    22.2 & 78.8 & 23.8&
    65.3 & 91.3 & 67.2\\
    Present (Transitive $\rightarrow$ Infinitive) &
    9.8 & 72.5 & 9.9&
    64.1 & 89.1 & 64.3\\
    Present (Complement) &
    18.6 & 82.8 & 20.2&
    79.9 & 96.0 & 80.6\\
    Present (Transitive $\rightarrow$ Complement) &
    53.4 & 88.9 & 57.3&
    81.9 & 96.5 & 84.7\\
    \midrule
    \multicolumn{7}{c}{Phrase Recombination}\\
    \midrule
    PP in Subj &
    11.9 & 71.4 & 14.2 &
    93.6 & 98.5 & 94.5\\
    PP in Indirect Obj &
    18.9 & 72.3 & 19.8 &
    92.5 & 98.1 & 93.6\\
    RC in Subj &
    1.2 & 58.5 & 2.2 &
    75.9 & 93.1 & 80.4\\
    RC in Indirect Obj &
    3.7 & 63.4 & 5.3 &
    54.9 & 87.9 & 57.5\\
    Adj in Subj &
    68.4 & 89.0 & 69.6 &
    96.5 & 99.1 & 97.1\\
    Adj in Indirect Obj &
    48.8 & 85.1 & 55.8 &
    98.8 & 99.7 & 99.3\\
    \midrule
    \multicolumn{7}{c}{Recursion Depth Alternation}\\
    \midrule
    CP recursion shallower &
    79.8 & 96.7 & ---&
    97.4 & 99.7 & ---\\
    CP recursion deeper &
    6.2 & 80.0 & ---&
    82.0 & 98.0 & ---\\
    PP recursion shallower &
    71.5 & 94.3 & 72.5&
    95.8 & 99.4 & 96.5\\
    PP recursion deeper &
    10.5 & 77.8 & 10.9&
    81.8 & 97.5 & 82.9\\
    CE recursion shallower &
    73.3 & 95.2 & ---&
    96.0 & 99.3 & ---\\
    CE recursion deeper &
    11.7 & 78.3 & ---&
    75.9 & 95.9 & ---\\
    Adj recursion shallower &
    99.2 & 99.7 & 99.4 &
    98.7 & 99.7 & 99.3\\
    Adj recursion deeper &
    99.5 & 99.8 & 99.8 &
    98.4 & 99.7 & 99.0\\
    \midrule
    \multicolumn{7}{c}{Gap Position Recombination}\\
    \midrule
    Indirect Obj-extracted RC &
    43.2 & 87.7 & 43.5&
    55.0 & 91.2 & 55.6\\
    Indirect Obj-extracted \textit{wh}-question &
    73.8 & 91.3 & ---&
    81.7 & 94.2 & ---\\
    \midrule
    \multicolumn{7}{c}{\textit{Wh}-question Structural Alternation}\\
    \midrule
    Active Subject &
    83.4 & 93.7 & ---&
    86.0 & 95.4 & ---\\
    Passive Subject &
    65.0 & 88.9 & ---&
    47.4 & 80.0 & ---\\
    Direct Obj, Ditransitive &
    84.7 & 95.3 & ---&
    51.0 & 79.8 & ---\\
    Subject with PP &
    0.1 & 65.6& ---&
    23.4 & 78.3 & ---\\
    Long Movement &
    19.8 & 77.0 & ---&
    3.6 & 63.3 & ---\\
    \bottomrule
    \end{tabular}
    \caption[Results by generalization pattern.]{Results by generalization pattern.}
    \label{table:result_pattern}
\end{table*}

Table~\ref{table:result_pattern} presents the evaluation results on each generalization pattern.

\subsection{Primitive Substitution}
Both Transformer and Llama~2 achieved more than 90\% Partial Match and more than 90 BLEU scores in the ``Subj $\rightarrow$ Obj'' and ``Obj $\rightarrow$ Subj'' patterns regardless of whether the target word was a proper noun or a common noun.
In contrast, the scores in the patterns regarding the generalizations from primitives were lower in both models.

The most frequent error made by Transformer in these patterns was to put an incorrect word whose first several characters are the same as those of the target word.
For example, in the ``Prim $\rightarrow$ Subj (proper)'' pattern, Transformer made the following mistake.
\begin{exe}
    \ex\begin{xlist}
        \exi{En:} Lina cooked the chicken.
        \exi{Ja-Gold:} \gll rina-ga tori-o ryourisi-ta\\
        {Lina-\Nom} {chicken-\Acc} {cook-\Pst}\\
        \exi{Ja-Pred:} \gll rinkaan-ga tori-o ryourisi-ta\\
        {Lincoln-\Nom} {chicken-\Acc} {cook-\Pst}\\
        \end{xlist}
\end{exe}
This error could have been due to the subword tokenization, so we tested Transformer trained without subword tokenization, but then it translated only the target word and not the whole sentence.
Therefore, we assume that these patterns are challenging for Transformer.
These results suggest that both models can generalize to unseen combinations of primitives seen in sentences and grammatical roles but struggle with generalizations using primitives not seen in sentences.

In addition, when comparing the scores between the patterns targeting common and proper nouns, the scores of the patterns targeting proper nouns were higher in most cases.
This result is opposite to that of \citet{kim-linzen-2020-cogs}, and the reason may be the difference in output formats between semantic parsing and machine translation, given that semantic parsing requires different representations for common and proper nouns.
\citet{kim-linzen-2020-cogs} mentioned that minor differences among lexical items may affect a model's performance, which can also explain the present results.

\subsection{Primitive Structural Alternation}
In the two patterns not requiring morphological generalization, namely, ``Double Obj $\rightarrow$ PP'' and ``PP $\rightarrow$ Double Obj'', Transformer and Llama~2 achieved near-perfect scores in Exact Match and BLEU.
This shows that both models can correctly translate sentences with different orders of arguments.

The patterns requiring morphological generalization such as ``Active $\rightarrow$ Passive'' were more challenging for both models.
Llama~2 achieved higher scores than Transformer, and we assume that this was because of pretraining without any control of the target verbs.

In the ``Active $\rightarrow$ Passive'' pattern, the errors made by Transformer are related to the passive form of a different verb, as in the following example.
\begin{exe}
    \ex\begin{xlist}
        \exi{En:} Sophia was recognized by Liam.
        \exi{Ja-Gold:} \gll sofia-ga riamu-niyotte ninsikisa-re-ta\\
        {Sophia-\Nom} {Liam-by} {recognize-\Pass-\Pst}\\
        \exi{Ja-Pred:} \gll sofia-ga riamu-niyotte sodate-rare-ta\\
        {Sophia-\Nom} {Liam-by} {raise-\Pass-\Pst}\\
        \end{xlist}
\end{exe}
This shows that Transformer did recognize the voice of the verb but failed to morphologically generalize to the passive form of the verb.
Similar errors were also made in other patterns in this category such as the ``Passive $\rightarrow$ Active'' pattern.
These results suggest that Transformer can generalize to the alternation of verb argument structure but struggles with morphological generalization in our settings.

\subsection{Tense Alternation}
Both Transformer and Llama~2 achieved higher scores in the patterns where the translation of the present form was given in the training set, except those regarding infinitives.
Generalization patterns regarding taking infinitives as the object were challenging for both models, especially for Transformer.
Considering that Llama~2 had already seen the target verbs in pretraining, its scores were low in these patterns.

Unsurprisingly, in the patterns requiring morphological generalizations to an unseen tense form of a familiar verb, none of the irregular verbs were translated correctly by Transformer.
In the ``Present (Ditransitive)'' pattern, regular verbs were translated correctly in about half of the cases.

On the other hand, in the ``Present (Infinitive)'' and ``Present (Complement)'' patterns, even regular verbs were translated incorrectly in most cases.
The difference between those patterns is that the latter involves function words such as \textit{to} and \textit{that}.
Also, the ``Present (Transitive) $\rightarrow$ Present (Infinitive)'' pattern was much more challenging for Transformer than the ``Present (Transitive) $\rightarrow$ Present (Complement)'' pattern.
This means that Transformer struggled in translating the combination of a verb in the present tense form and \textit{to} (e.g., plans to) even though the combination of the verb in the past tense form and \textit{to} (e.g., planned to) was seen in the training set.
These results suggest that Transformer encodes a verb and \textit{to} together, which makes it difficult to generalize to a different tense form of the verb.
This can be considered as Transformer overfitting the training set and local syntactic structures therein.

As for the patterns regarding complement clauses, because \textit{that} is used in two ways in the training set (as a complementizer and as a relative pronoun), Transformer seems to have struggled to distinguish them when required to generalize to the unseen tense form of a verb and \textit{that} (e.g., 
\textit{realizes that}, when only \textit{realized} is included in the training set).
However, it can generalize to the unseen combination of a familiar verb and \textit{that} (e.g., \textit{understands that}, when \textit{understands} is included in the training set) because it can rule out the possibility of \textit{that} being a relative pronoun.

\subsection{Phrase Recombination}
\begin{table*}[t]
    \centering
        \small
        \begin{tabular}{lcccccc}
            \toprule
            & \multicolumn{3}{c}{Transformer} & \multicolumn{3}{c}{{Llama~2}}\\
            Generalization pattern & Exact & BLEU & Partial & Exact & BLEU & Partial\\
            \midrule
            PP in Subj not in CP &
            3.5 & 66.6 & 4.0 &
            94.8 & 98.8 & 95.4\\
            PP in Subj in CP &
            20.2 & 76.2 & 24.5 &
            92.5 & 98.1 & 93.6\\
            PP in Indirect Obj not in CP &
            23.4 & 69.5 & 24.6 &
            92.6 & 97.7 & 93.3\\
            PP in Indirect Obj in CP &
            14.4 & 75.1 & 15.0 &
            92.4 & 98.4 & 93.9\\
            RC in Subj not in CP &
            0.3 & 52.2 & 1.0 &
            75.8 & 92.8 & 80.0\\
            RC in Subj in CP &
            2.2 & 64.7 & 3.3 &
            76.0 & 93.4 & 80.9 \\
            RC in Indirect Obj not in CP &
            6.0 & 61.6 & 8.4 &
            55.0 & 86.1 & 56.8\\
            RC in Indirect Obj in CP &
            1.3 & 65.3 & 2.1 &
            54.8 & 89.8 & 58.2\\
            Adj in Subj not in CP &
            65.4 & 86.1 & 66.2 &
            94.7 & 98.5 & 95.0 \\
            Adj in Subj in CP &
            71.3 & 91.9 & 72.9 &
            98.3 & 99.7 & 99.2\\
            Adj in Indirect Obj not in CP &
            54.3 & 83.8 & 60.1 &
            99.4 & 99.8 & 99.6\\
            Adj in Indirect Obj in CP &
            43.3 & 86.4 & 51.6 &
            98.1 & 99.7 & 99.0\\
            \bottomrule
        \end{tabular}
        \caption[Comparison of results with/without modified phrases in CP.]{Comparison of results with/without modified phrases in CP.}
        \label{table:mod_result}
\end{table*}

Transformer struggled with the patterns in this category, especially those involving PPs and RCs.
Its scores in the patterns involving PPs and RCs in the indirect object position were higher than those when in the subject position, but they were still low.
Both Transformer and Llama~2 achieved higher scores in the patterns involving PPs than in those involving RCs, which is consistent with the results of \citet{li-etal-2023-slog}, and achieved much higher scores in the patterns with the adjectives in an unseen position.
The reason is that it is easier to translate sentences whose dependencies between the noun and the modifier are shorter, as pointed out by \citet{li-etal-2023-slog}.
Also, Transformer performed better with adjectives in the subject than with those in the indirect position.

In addition, the scores were influenced greatly by whether or not RCs or PPs in an unseen position are in a complement clause.
Table~\ref{table:mod_result} compares the scores between the two settings.
It shows that Transformer generalized better to a modified phrase in a subject position when it was in a complement clause than when it was not, and that the opposite was true for a modified phrase in an indirect object position.

Common errors made by Transformer came from ignoring the modifiers in translation or considering the modified phrases as the direct object.
For example, in the ``PP in Subject'' pattern, Transformer made the following mistakes.
\begin{exe}
    \ex\begin{xlist}
        \exi{En:} A friend in the house was given the book.
        \exi{Ja-Gold:} \gll ie-no naka-no tomodati-ga hon-o age-rare-ta\\
        {house-\Gen} {in-\Gen} {friend-\Nom} {book-\Acc} {give-\Pass-\Pst}\\
        \exi{Ja-Pred:} \gll tomodati-ga hon-o age-rare-ta\\
        {friend-\Nom} {book-\Acc} {give-\Pass-\Pst}\\
        \end{xlist}
    \ex\begin{xlist}
        \exi{En:} A jar on the book changed.
        \exi{Ja-Gold:} \gll hon-no ue-no bin-ga kawat-ta\\
        {book-\Gen} {on-\Gen} {jar-\Nom} {change-\Pst}\\
        \exi{Ja-Pred:} \gll bin-ga hon-no ue-no bin-o kae-ta\\
        {jar-\Nom} {book-\Gen} {on-\Gen} {jar-\Acc} {change-\Pst}\\
        \end{xlist}
\end{exe}
Such errors occurred regardless of whether or not the modified phrase was in a complement clause, but they were more frequent in sentences where the modified phrase was not in a complement clause but in the beginning of the sentence.
This suggests that Transformer encodes the first noun of the sentence as the subject without considering its modifiers and encodes the subject in a complement clause differently.
When the modified phrase in the subject position is in a complement clause, the model can translate it correctly more often because it is not the first noun in the sentence.

Also, \citet{li-etal-2023-slog} mentioned that in the ``PP in Subj not in CP'' pattern, Transformer tended to take the nearest noun from the verb as the subject in SLOG, whereas it tended to take the first noun in the sentence as the subject in our experiment.
One explanation for these errors is that Transformer learned only the local syntactic structures, which is not sufficiently robust for generalizing to unseen structures.

\subsection{Recursion Depth Alternation}
Transformer struggled with deeper recursions except for adjective recursions, although we added concatenated sentences to the training set to remove the need for length generalization.
\citet{li-etal-2023-slog} focused on generalization sentences that are longer than any sentences in the training set and showed that Transformer still struggled without being required to generalize to longer sentences.
However, that is limited in that it excludes longer sentences that may be more difficult to translate, which may affect the model's performance.
Our method does not exclude any sentences in the generalization set, and it is more thorough than the existing methods.
Nevertheless, the present finding above regarding Transformer struggling is still consistent with that reported by \citet{li-etal-2023-slog}.

On the other hand, Transformer achieved near-perfect scores in deeper adjective recursions.
We assume that this is because the length of recursions is shorter in adjective recursions, and the order of the adjective words is the same in English and Japanese.
These results indicate that Transformer generally struggles with longer recursions, but it can generalize to deeper recursions whose structure is simple.

Regardless of the type of recursion and whether the novel depth is deeper or shallower, one of the recursions is skipped in most of the errors made by Transformer.
For example, in the ``PP recursion shallower'' pattern, Transformer made the following mistake.
\begin{exe}
    \ex\begin{xlist}
        \exi{En:} The child handed the box beside a table beside a tree beside a house to the teacher.
        \exi{Ja-Gold:} \gll kodomo-ga ie-no yoko-no ki-no yoko-no teeburu-no yoko-no hako-o kyoosi-ni tewatasi-ta\\
        {child-\Nom} {house-\Gen} {side-\Gen} {tree-\Gen} {side-\Gen} {table-\Gen} {side-\Gen} {box-\Acc} {teacher-\Dat} {hand-\Pst}\\
        \exi{Ja-Pred:} \gll kodomo-ga ie-no yoko-no ki-no yoko-no hako-o kyoosi-ni tewatasi-ta\\
        {child-\Nom} {house-\Gen} {side-\Gen} {tree-\Gen} {side-\Gen} {box-\Acc} {teacher-\Dat} {hand-\Pst}\\
        \end{xlist}
\end{exe}
We argue that this error occurred because the model struggled with unseen long-distance dependencies.

\subsection{\textit{Wh}-question Structural Alternation}
Transformer achieved higher scores in the ``Active Subject'' pattern than in the ``Passive Subject'' pattern, which is consistent with the results of \citet{li-etal-2023-slog}.
However, the performance of Transformer and Llama~2 in translating \textit{wh}-questions with the passive voice was about 30\% better than that in SLOG.
These results suggest that \textit{wh}-questions with the passive voice are challenging in semantic parsing, but not necessarily in structural generalization in general.

Many of the errors made by Transformer came from mistakes about the grammatical role of the interrogative pronoun.
For example, in the ``Passive Subject'' pattern, Transformer made the following mistakes.
\begin{exe}
    \ex\begin{xlist}
        \exi{En:} What was seen?
        \exi{Ja-Gold:} \gll nani-ga mi-rare-ta-ka?\\
        {what-\Nom} {see-\Pass-\Pst-\Qp}\\
        \exi{Ja-Pred:} \gll nani-ga nani-o mi-rare-ta-ka?\\
        {what-\Nom} {what-\Acc} {see-\Pass-\Pst-\Qp}\\
        \end{xlist}
    \ex\begin{xlist}
        \exi{En:} What was brought to the boy?
        \exi{Ja-Gold:} \gll nani-ga syoonen-ni motteko-rare-ta-ka?\\
        {what-\Nom} {boy-\Dat} {bring-\Pass-\Pst-\Qp}\\
        \exi{Ja-Pred:} \gll nani-ga nani-ni motteko-rare-ta-ka?\\
        {what-\Nom} {what-\Dat} {bring-\Pass-\Pst-\Qp}\\
        \end{xlist}
\end{exe}
Transformer added an interrogative pronoun as the direct object or the indirect object, while it should be the subject of the verb.
This error may be because \textit{wh}-questions are encoded in a way that \textit{what} is the object of their verb because \textit{what} is always a direct object in the training set.
This means that the model could not generalize to the unseen grammatical role of the interrogative pronoun because of the structurally incorrect encoding of \textit{wh}-questions.

In the ``Direct Obj, Ditransitive'' pattern, Transformer achieved more than 80\% Exact Match and 90 BLEU scores, although it scored lower than 20\% Exact Match in the same pattern in SLOG.
Surprisingly, Llama~2 performed worse than Transformer in this pattern.
The errors made exclusively by Llama~2 in this pattern contained structures that were not combinations of any familiar ones in the training set, thus suggesting an influence of pretraining.
Finally, both models struggled with unseen long-distance dependencies in the ``Subject with PP'' and ``Long movement'' patterns, which is consistent with the results of \citet{li-etal-2023-slog}.

\section{Ethical Considerations}
This paper is focused on creating a benchmark for evaluating compositional generalization in machine translation and using it to analyze neural models.
It does not include any contents that can be potentially used for harmful applications.
Also, the benchmark is generated with PCFGs, and it contains neither offensive content nor any information that could be used to identify individual people.

\end{document}